\documentclass[twoside]{article}

\usepackage{PRIMEarxiv}

\usepackage[utf8]{inputenc} 
\usepackage[T1]{fontenc}    
\usepackage{hyperref}       
\usepackage{url}            
\usepackage{booktabs}       
\usepackage{amsfonts}       
\usepackage{nicefrac}       
\usepackage{microtype}      
\usepackage{lipsum}
\usepackage{fancyhdr}       
\usepackage{graphicx}       
\graphicspath{{media/}}     

\usepackage{graphicx}%
\usepackage{multirow}%
\usepackage{amsmath,amssymb,amsfonts}%
\usepackage{amsthm}%
\usepackage{mathrsfs}%
\usepackage[title]{appendix}%
\usepackage{xcolor}%
\usepackage{textcomp}%
\usepackage{manyfoot}%
\usepackage{booktabs}%
\usepackage{algorithm}%
\usepackage{algorithmicx}%
\usepackage{algpseudocode}%
\usepackage{listings}%

\usepackage{amsmath} 

\usepackage{float} 
\usepackage{booktabs}
\usepackage{array}
\usepackage{tabularx}
\usepackage{float} 
\usepackage{caption} 
\usepackage{subcaption}

\title{Large Language Models: A New Approach for Privacy Policy Analysis at Scale
}

\author{
  David Rodriguez\\
  ETSI Telecomunicación\\
  Universidad Politécnica de Madrid\\
  Spain\\
  \texttt{david.rtorrado@upm.es} \\
  \And
  Ian Yang\\
  School of Computer Science\\
  Carnegie Mellon University\\
  Forbes Ave, Pittsburgh, Pennsylvania\\
  United States\\
  \texttt{iyang30@gatech.edu} \\
  \And
  Jose M. Del Alamo\\
  ETSI Telecomunicación\\
  Universidad Politécnica de Madrid\\
  Spain\\
  \texttt{jm.delalamo@upm.es} \\
  \And
  Norman Sadeh\\
  School of Computer Science\\
  Carnegie Mellon University\\
  Forbes Ave, Pittsburgh, Pennsylvania\\
  United States\\
  \texttt{sadeh@cs.cmu.edu}
}
\begin{document}
\maketitle

\begin{abstract}
    The number and dynamic nature of web and mobile applications presents significant challenges for assessing their compliance with data protection laws. In this context, symbolic and statistical Natural Language Processing (NLP) techniques have been employed for the automated analysis of these systems’ privacy policies. However, these techniques typically require labor-intensive and potentially error-prone manually annotated datasets for training and validation. This research proposes the application of Large Language Models (LLMs) as an alternative for effectively and efficiently extracting privacy practices from privacy policies at scale. Particularly, we leverage well-known LLMs such as ChatGPT and Llama 2, and offer guidance on the optimal design of prompts, parameters, and models, incorporating advanced strategies such as few-shot learning. We further illustrate its capability to detect detailed and varied privacy practices accurately. Using several renowned datasets in the domain as a benchmark, our evaluation validates its exceptional performance, achieving an F1 score exceeding 93\%. Besides, it does so with reduced costs, faster processing times, and fewer technical knowledge requirements. Consequently, we advocate for LLM-based solutions as a sound alternative to traditional NLP techniques for the automated analysis of privacy policies at scale.
\end{abstract}


\section{Introduction}\label{Introduction}

The digital era has led to an unprecedented expansion of web and mobile applications and a myriad of online services. This growth is a testament to technological advancement and the increasing reliance of businesses and organizations on digital platforms for various operations. A central aspect of this digital proliferation is the extensive use of technologies for personal data collection, primarily driven by the objective of enhancing targeted marketing strategies. The ability to collect, analyze, and utilize user data has become a cornerstone of modern commerce, offering businesses invaluable insights into consumer behavior and preferences.

However, the increasing collection and utilization of personal data have raised significant privacy concerns. Users' privacy is at risk as their data becomes valuable in the digital marketplace. This concern has led to the emergence of regulatory bodies and the formulation of data protection legislation aimed at safeguarding user privacy. These legislations, such as the General Data Protection Regulation (GDPR) in the European Union and the California Consumer Privacy Act (CCPA) in the United States, impose stringent requirements on how organizations should handle personal data.

Ensuring compliance with these legislations, however, poses a formidable challenge. The overgrowth of online services, compounded by globalization, makes it impractical, if not impossible, for regulators to manually assess each service’s adherence to privacy laws. This situation is further exacerbated by the dynamic nature of online services, where data processing practices and the privacy policies disclosing them are subject to frequent changes \cite{srinath2023privacy}. In response to these challenges, automated-driven methods have been proposed for analyzing privacy policies at scale \cite{delalamo2022systematic}. This approach holds the potential for research endeavors aimed at understanding and highlighting privacy concerns, and it also offers a practical tool for regulators to conduct mass evaluations of applications and services, thereby promoting a higher level of compliance with privacy regulations.

The automated analysis of privacy policies has leveraged Natural Language Processing (NLP) techniques \cite{delalamo2022systematic}. Symbolic and statistical state-of-the-art NLP techniques are proposed to address this task, although each has drawbacks. Symbolic approaches rely on pre-defined rules, leading to lower performance when compared to statistical approaches due to the lack of adaptability to differences present in legal texts. Thus, state-of-the-art research has predominantly relied on statistical approaches such as machine learning (ML) techniques, and particularly supervised learning models, to train and evaluate models identifying privacy practice disclosures such as personal data collection or sharing \cite{zimmeck2017automated}.  These models require the use of manually annotated datasets \cite{wilson2018analyzing, bui2021automated, harkous2018polisis}, which are often expensive, time-consuming to create, and prone to errors \cite{klie2023annotation}. Furthermore, building and training those models require advanced technical expertise, contributing to a higher barrier to entry. As a result, their practical application is mainly suitable for large-scale projects where the benefits can outweigh these significant costs. On the other hand, modern Generative Artificial Intelligence (GenAI), particularly Large Language Models (LLMs), represents a significant advancement in the NLP domain, being able to understand and generate human-like text, making it particularly well-suited for parsing and analyzing the complex language present in privacy policies without needing annotated datasets. In this context, this paper proposes the application of LLMs for the effective and efficient extraction of privacy practices from privacy policies. In particular, we focus on ChatGPT, which relies on Generative Pre-trained Transformer (GPT) models. 

Our study identifies the optimal configuration of ChatGPT prompts, parameters, and models, integrating advanced techniques such as few-shot learning. Additionally, we conduct a comparative analysis of our proposed ChatGPT configuration with Llama 2 and other state-of-the-art techniques. Our findings reveal that our proposal competes with and even outperforms these traditional methods. Moreover, we discuss its advantages regarding lower upfront costs, reduced processing times, and greater ease of use.

Thus, we propose LLM-based solutions and our specific ChatGPT configuration as a viable replacement for traditional NLP techniques in the task of automated privacy policy processing. Our research contributes to the ongoing discourse on the application of GenAI in legal and regulatory contexts \cite{choi2023chatgpt, tan2023chatgpt, tang2023policygpt}, suggesting a paradigm shift towards more efficient, cost-effective, and accessible tools for privacy policy analysis.

\section{Related Work}\label{Related Work}

Privacy policies are documents written in plain text that outline how organizations handle personal data. However, the complexity and length of these documents often make them challenging to understand and process \cite{reidenberg2015disagreeable}. This has spurred interest in automated methods for analyzing privacy policies \cite{delalamo2022systematic}, which fall into two major categories, namely, symbolic and statistical NLP. 

Symbolic NLP approaches \cite{dAquin2018PrivOnto, Evans2017HyponymyExtraction, Hosseini2016LexicalPrivacyPolicies} are relevant but come with inherent limitations when processing new texts: these techniques model language through grammar rules and lexicons, thus requiring extensive manual effort to create and code these rules. This process is both time-consuming and hard to scale, especially when dealing with intricate aspects of privacy policies. Symbolic NLP is effective in morphological and lexical analysis, such as identifying privacy practices through keyword analysis. It also handles more complex tasks like syntactic and semantic analysis, using tools like the Stanford dependency parser \cite{Chen2014DependencyParser}. PolicyLint \cite{Andow2019PolicyLint} is a state-of-the-art tool based on this symbolic NLP approach that employs ontologies to detect contradictions in privacy policy statements about personal data collection and sharing. Its ability to identify negative sentences —a challenging task for statistical NLP techniques— highlights its potential for specific privacy policy analysis tasks. However, it faces challenges with unanticipated variations, including typos or infrequent cases, thus limiting its applicability to new cases. 

Statistical NLP approaches, on the other hand, leverage machine learning techniques for language processing: supervised, unsupervised, and Artificial Neural Networks (ANN)-based techniques. Supervised methods are the predominant technique usually employed for automated privacy policy analysis, with geometric algorithms like Support Vector Machine (SVM) \cite{Guntamukkala2015PrivacyPolicies, wilson2016websitePrivacyPolicy, daSilva2016PrivacyPolicy, story2019nlpPrivacy} and Logistic Regression (LR) \cite{sathyendra2016optOutExtraction, sathyendra2017identifying} being the most prevalent. Unsupervised techniques, although less common, utilize models like Hidden Markov Models (HMM) \cite{liu2014privacyAlignment, wilson2016websitePrivacyPolicy} and Latent Dirichlet Allocation (LDA) \cite{massey2013automatedPolicyAnalysis} for clustering practices during policy analysis. ANN-based techniques are also in use for this task, including Convolutional Neural Networks (CNNs) \cite{harkous2018polisis, keymanesh2020toward}, Recurrent Neural Networks (RNN) \cite{liu2016modeling}, and Google’s BERT \cite{wilson2018analyzing, ravichander2019question}, sometimes showing superior performance than supervised learning methods \cite{wilson2018analyzing, ravichander2019question}.

The development of new privacy policy analysis methods leveraging statistical NLP approaches frequently requires labeled corpora for training and validation. In the domain of privacy policy analysis, several datasets manually annotated by legal experts have been employed to build supervised learning methods. MAPP corpus \cite{arora2022privacyPolicyCorpus} is a public multilingual dataset that contains 64 Google Play Store app privacy policies, chunked into segments (i.e., paragraphs) manually annotated as disclosing the collection or sharing of various types of personal data. The OPP-115 dataset \cite{wilson2016websitePrivacyPolicy} is the most widely used dataset in the domain. It follows the same structure and contains annotations of almost identical practices and data types to MAPP but with a larger number (n=115) of annotated policies. APP-350 \cite{zimmeck2019maps} is the largest dataset of this type, which has 350 privacy policies with annotations of collection/sharing of even more specific data types than the previous ones. Likewise, the IT100 \cite{IT100Corpus} is a dataset of 100 privacy policies containing annotations of statements disclosing international transfers of personal data.

Expanding upon symbolic and statistical NLP methods, LLMs can generate coherent text based on a given input, such as GPTs \cite{radford2018improving} and Llama 2 \cite{touvron2023llama}. Building on those LLMs, ChatGPT and Llama 2-Chat are chatbots trained to provide meaningful answers to pieces of text inputs (i.e., prompts) and with adjustable performance through parameters like “temperature” that influence the results’ variability \cite{ghanadian2023chatgpt}, and response times. The ability to provide relevant answers is achieved through a combination of unsupervised and supervised learning techniques underpinned by neural networks trained on extensive datasets. Additionally, the relevance and format of the responses are typically enhanced through prompt augmentation \cite{shum2023automatic}, which involves modifying the given input prompt to improve the output performance or to steer the output in a specific direction. Notably, LLMs’ proficiency in processing lengthy input texts is boosted by the attention mechanisms inherent in transformer architectures \cite{vaswani2017attention}. A recent study conducted by Qin et al. \cite{qin2023chatgpt} has analyzed to what extent LLMs like ChatGPT can perform various tasks —reasoning, language inference, Q\&A, dialogue, summarization, entity recognition, and sentiment analysis— using 20 well-established NLP datasets to benchmark their performance, showing high reasoning capabilities.

Integrating LLMs into the automated analysis of privacy policies and legal texts \cite{savelka2023unreasonable} represents a significant evolution in assessing compliance with data protection regulations. Tang et al. \cite{tang2023policygpt} have explored their application in this context, highlighting its potential to surpass traditional methods in extracting and classifying general, coarse-grained privacy practices within legal texts. Our research extends this exploration by thoroughly analyzing LLMs’ ability to identify more detailed data practices in privacy policies. We provide insights into the optimal model configuration for this task and further demonstrate LLMs’ generalization capabilities, particularly ChatGPT’s, to identify varied privacy practices. Our findings reveal that ChatGPT, leveraging few-shot learning, outperforms traditional symbolic and statistical NLP methods in key areas, including classification performance, time efficiency, and cost-effectiveness.

\section{Experimental Design}\label{Experimental Design}

GPT models have an intrinsically complex behavior dependent on the prompt design, configuration parameters, and model selection. We rely on the Design Science Research (DSR) methodology \cite{vomBrocke2020design} to propose a ChatGPT framework for privacy policy analysis and evaluate its effectiveness. DSR guides the design of new artifacts through an iterative and systematic process. Specifically, we followed an iterative split testing process \cite{kohavi2015online} to assess the performance of the prompt, parameter, or model selection changes within each iteration. Finally, we check our proposed configuration performance against two unseen sets of policies, conduct a set of comparative analyses with state-of-the-art solutions, and demonstrate its generalization capabilities. Through this systematic process, we propose a well-performing and generalizable configuration of ChatGPT as a novel and effective approach for the automated analysis of privacy policies. 

\subsection{Ground Truth}\label{Ground Truth}

Determining the optimal ChatGPT configuration that offers the best performance requires using a ground truth dataset to validate and quantify results. We relied on the MAPP dataset \cite{arora2022privacyPolicyCorpus} for this task, retaining an experimental set on which to apply changes and measure their impacts and a control set to validate the final configuration's overall performance. Unlike traditional NLP techniques, using a ground truth dataset is only required while designing the configuration framework. Afterward, we can generalize it to identify other privacy practices without generating new annotated datasets or validating new methods, as demonstrated in Section \ref{Demonstration}.

The MAPP dataset is unbalanced. Thus, we stratified sampling to generate the experimental (33 policies) and control (31 policies) subsets. With a standard deviation between both datasets of 2.44 for all data categories, with categories annotated in almost all policies (e.g., IP address and device IDs, in 95\%) and others in only one of them (e.g., Political, religious, or philosophical belief). This is contextualized against the backdrop of the mean annotation counts per policy, which are 8.13 and 8.18 for the experimental and control sets, respectively. The observed standard deviation, in relation to the means, suggests a moderate degree of variance in annotation frequency per policy across the datasets. This degree of variability is within acceptable limits for the intended analytical scope, affirming a balanced and representative data stratification for the empirical analysis.

\subsection{Prompt Design}

\begin{figure}[h]
  \centering
  \begin{subfigure}[t]{0.48\textwidth}
    \centering
    \includegraphics[width=\linewidth]{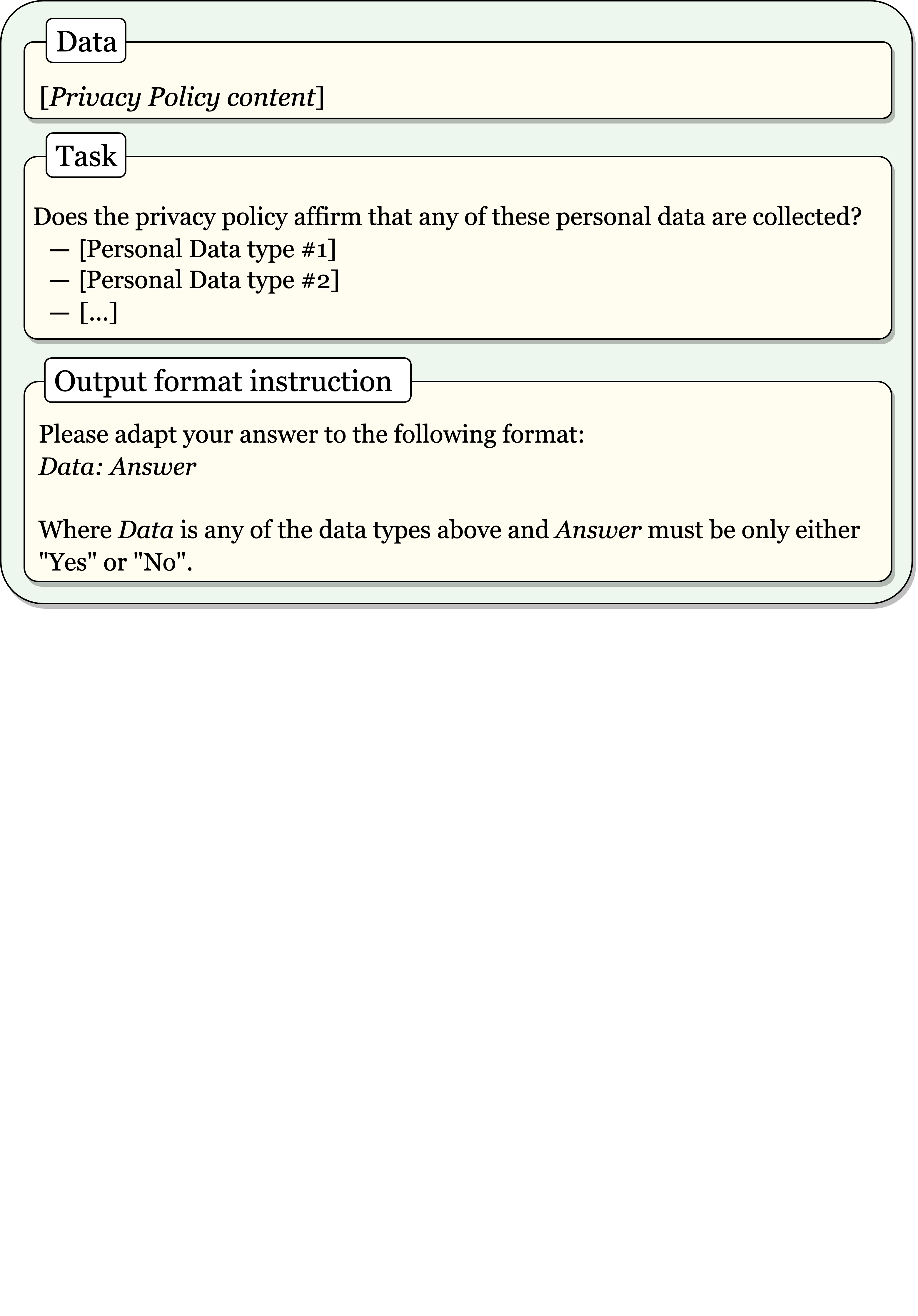}
    \caption{Baseline prompt design.}
    \label{fig:Baseline_prompt}
  \end{subfigure}
  \hfill
  \begin{subfigure}[t]{0.48\textwidth}
    \centering
    \includegraphics[width=\linewidth]{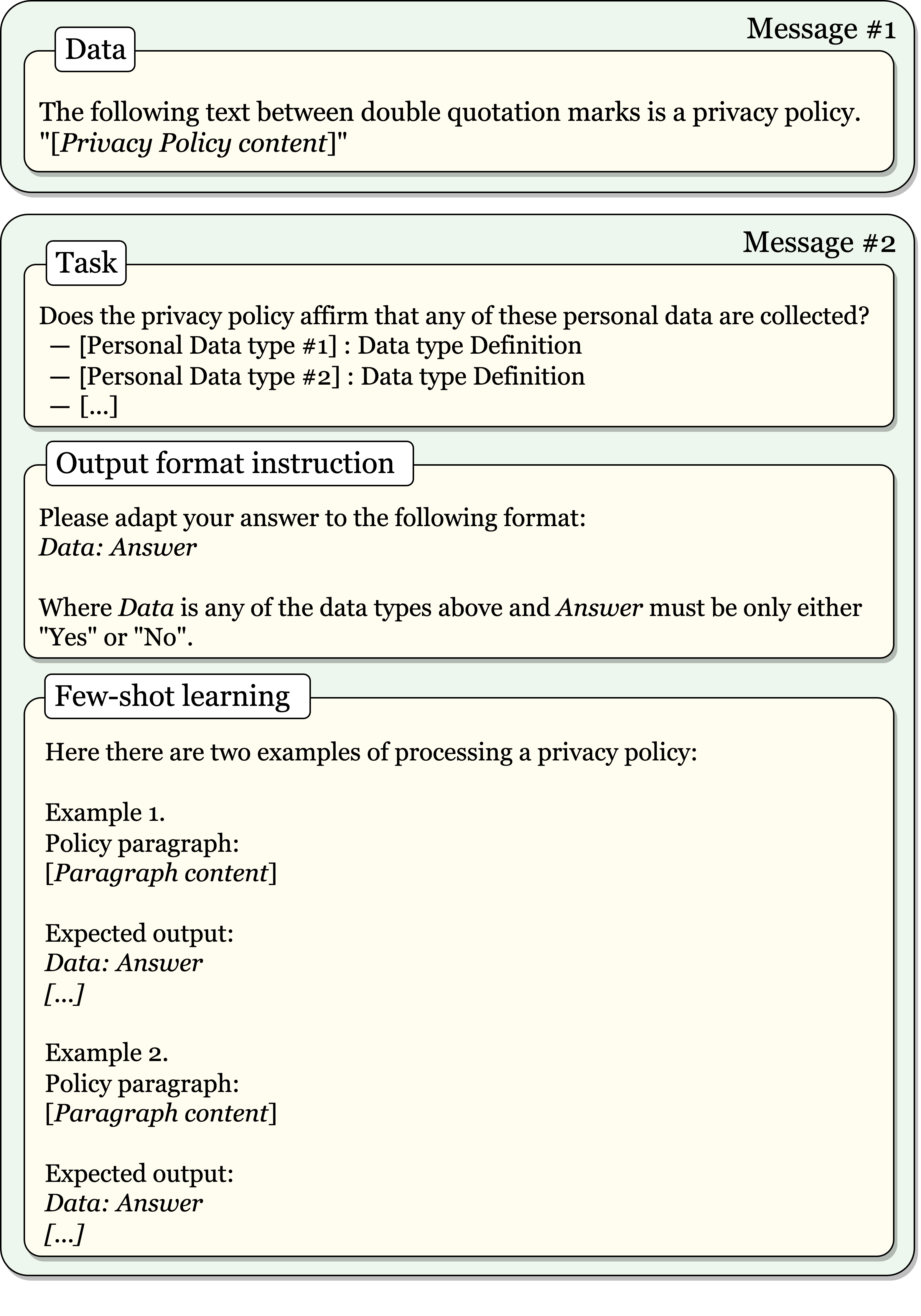}
    \caption{Final prompt design.}
    \label{fig:Enhanced_prompt}
  \end{subfigure}
  \caption{Baseline and final ChatGPT prompt designs for the identification of privacy practices in privacy policies.}
  \label{fig:Prompt_design_comparison}
\end{figure}

We departed from the prompt design depicted in Figure \ref{fig:Baseline_prompt}. This baseline prompt is structured in \emph{Data}, \emph{Task} and \emph{Output} format instruction segments. \emph{Data} is the privacy policy for identifying practices. \emph{Task} is the actual practice in which identification in the policy is requested. \emph{Output} format instruction provides the guidelines to obtain responses that can be processed automatically. 

We applied this baseline prompt to the specific task of identifying types of personal data purportedly collected or shared as per the privacy policy. This required providing the privacy policy in the Data segment, asking about each data type in the Task segment, and steering the formatting of responses, including “\emph{Data: Answer}” in the Output format instruction segment. In our experiment, this baseline prompt achieved the following metrics: 0.79 accuracy, 0.92 recall, 0.78 precision and 0.84 F1 score. 

Crafting the optimum prompt design requires a split testing process to reveal the effects of various changes in the prompt. The only ChatGPT parameter adjusted during this phase is the temperature value, set to zero to provide more deterministic responses \cite{OpenAIAPI}, and using the GPT-4 Turbo model to take advantage of its speed and input prompt size capabilities up to 128k tokens. While numerous tests were conducted, this section will focus solely on those that involved a significant change in performance metrics. A detailed summary of these metrics, derived from each test, is encapsulated in Table \ref{tab:summary_tests}. In this testing sequence, each technique that demonstrated a performance improvement was systematically integrated into the subsequent tests. Thus, each new test was benchmarked against the last updated configuration.

\textbf{Specifying Data boundaries.} Incorporating the phrase “The following text is a privacy policy” improved the metrics, specifically a +1.47\% increase in accuracy (from 0.793 to 0.804), +0.4\% in recall, +1.36\% in precision, and +0.92\% in F1 score. This improvement is attributed to enhancing the model's ability to discern the limits of the privacy policy text. A minor adjustment involving the indication that the privacy policy text is enclosed in double quotes led to an additional +0.8\% rise in recall, precision, and F1 score, and a +1.16\% increase in accuracy.

\textbf{Data placement.} We evaluated the impact of the placement of the privacy policy within the prompt —either at the beginning or the end— on the performance metrics. Positioning the privacy policy at the end, contrary to the beginning, slightly diminished the overall metrics, including a decrease in accuracy by -1.15\%, recall by -1.98\%, precision by -0.07\%, and F1 score by -0.97\%.

\textbf{Augmenting Task description}. The initial prompt version primarily focused on enumerating the types of data to be identified. However, given the inherent complexity of data categorization —a challenge even for human annotators as substantiated in related literature \cite{fredriksson2020}— the prompt was augmented to include the internal definitions used for manual annotations in the MAPP dataset. While this expansion resulted in a lengthier prompt, it significantly enhanced all metrics except for recall, with an increase of +4.58\% in accuracy, a decrease of -0.79\% in recall, +5.96\% in precision, and +2.72\% in F1 score.

\textbf{Message splitting.} We have tested splitting the prompt into two different messages, passed to ChatGPT one after the other. Specifically, we separated the privacy policy (Data segment) from the remainder prompt. The results show a better overall understanding and comprehension of the privacy policy, reflected in a +1.56\% increase in precision, +0.56\% increase in accuracy and +0.22\% increase in F1 score, at the cost of a -1.2\% decrease in recall. 

\textbf{Data pruning.} This technique eliminates the paragraphs of the policy that do not have information regarding collecting or sharing personal data. We crafted a specific prompt for this task. The results show the overall policy metrics have remained practically the same. 

\textbf{Segmentation.} We also assessed the role of input processing in the results with three different configurations: 1) Data segmentation, i.e., analyzing each individual paragraph at a time; 2) Task segmentation, i.e., asking only for one specific practice (e.g., a given data type collection) at a time, and 3) Data and Task segmentation, i.e., asking for one specific practice in one specific paragraph. Data segmentation did not show significant improvement. However, Task segmentation had a surprising effect on the result: accuracy decreased by -5.48\%, recall decreased by -18.4\%, precision increased by +8.06\%, and overall leading to a decrease in the F1 score of -6.46\%. This suggests that asking for each practice individually may lead to a loss of a broader contextual understanding of the model, negatively impacting its overall performance. Finally, Data and Task segmentation showed the worst results, dropping recall by -59.6\% and decreasing F1 score by -40.97\%, reinforcing the importance of context for ChatGPT when analyzing privacy policies.

It's worth noting that while Data segmentation showed a similar performance to keep the whole policy, it increases the cost of the queries since they are computed according to the prompt and response size and not specifically by the number of requests. Furthermore, it also increases the processing time for each policy, as more requests (as many as policy paragraphs) must be processed. Thus, we have discarded this option in favor of processing the whole policy.

\textbf{Few-shot prompting.} Few-shot prompting \cite{brown2020language} refers to providing a set of examples (shots) with the prompt to guide the model. We have tested this technique, including in the prompt one, two, and three examples —randomly chosen— of paragraph annotations. The best result was obtained with two-shot examples, showing a significant improvement in the metrics (+3.31\% accuracy, +0.0\% recall, +4.29\% precision, +2.18\% F1 score). 

\textbf{Final prompt design.} \label{Final_Prompt_Section} Figure \ref{fig:Enhanced_prompt} presents the prompt configuration that our tests have consistently found to be most effective in identifying privacy practices. In this optimized prompt structure, the Data segment is introduced in an initial message, followed by the Task segment in a subsequent message. This Task segment incorporates definitions of the targeted practice —in this instance, data types— along with the same Output format instruction used in the Baseline prompt and a Few-shot learning component. The Few-shot learning part includes two illustrative examples (Two-shot) of processing paragraphs from privacy policies and the corresponding expected outputs. 

\begin{table}[h]
\centering
\caption{Metrics obtained with each technique tested on ChatGPT and Llama 2. The tests have been executed sequentially and are incremental:  when a technique exhibits superior performance (highlighted in bold within the table), it is incorporated into the subsequent test. The sole exception to this is the Data Segmentation test, which demonstrated a minimal improvement in the F1 score at the expense of significantly increased processing time and a decrease in precision; hence, this technique has not been ultimately integrated. GPT-4 Turbo and Llama 2 70B-Chat models are used unless otherwise stated, as they showed the best performance.}
\label{tab:summary_tests}
\begin{tabular}{@{}llcccc@{}}
\toprule
\addlinespace[0.5em]
 & \textbf{Prompt technique} & \textbf{Accuracy} & \textbf{Precision} & \textbf{Recall} & \textbf{F1 Score} \\ 
\multirow{16}{*}{\rotatebox[origin=c]{90}{\textbf{ChatGPT}}} & & & & & \\
\cmidrule(lr){2-6} 
\addlinespace[0.5em] 
& \textbf{Baseline prompt}     & 0.793      & 0.785       & 0.922    & \textbf{0.848}      \\
& \textbf{Specify Data boundaries}    & 0.804      & 0.796       & 0.926    & \textbf{0.856}      \\
& \textbf{Specify Data boundaries (double quotes)}    & 0.814      & 0.803       & 0.933    & \textbf{0.863}      \\
& Data placement (Bottom)    & 0.804      & 0.802       & 0.915    & 0.855      \\
& \textbf{Augmenting Task description}     & 0.851      & 0.850       & 0.926    & \textbf{0.887}      \\
& \textbf{Message splitting}     & 0.855      & 0.864       & 0.915    & \textbf{0.888}      \\
& Data pruning     & 0.848      & 0.847       & 0.926    & 0.885      \\
& Data segmentation     & 0.850      & 0.837       & 0.948    & 0.889      \\
& Task segmentation     & 0.804      & 0.919       & 0.756    & 0.829      \\
& Data \& Task segmentation     & 0.571      & 0.871       & 0.374    & 0.523      \\
& \textbf{One-shot prompting}     & 0.862      & 0.889       & 0.893    & \textbf{0.891}      \\
& \textbf{Two-shot prompting}     & 0.874      & 0.886       & 0.919    & \textbf{0.902}      \\
& Three-shot prompting     & 0.858      & 0.872       & 0.907    & 0.889      \\
\addlinespace[0.5em]
\multirow{7}{*}{\rotatebox[origin=c]{90}{\textbf{Llama 2}}} & & & & & \\
& \textbf{Baseline prompt}     & 0.846      & 0.880       & 0.873    & \textbf{0.877}      \\
& Data segmentation     & 0.625      & 0.625       & 1.000    & 0.770      \\
& Task segmentation     & 0.613      & 0.870       & 0.490    & 0.627      \\
& \textbf{Data \& Task segmentation}     & 0.846      & 0.847       & 0.921    & \textbf{0.882}      \\
& Two-shot prompting     & 0.623      & 0.623       & 1.000    & 0.768      \\
\addlinespace[0.5em] 
\bottomrule
\end{tabular}
\end{table}

\subsection{Parameter Tuning}\label{Parameter Tuning}

ChatGPT offers a number of parameters that can be configured to modify its responses \cite{OpenAIAPI}, namely temperature, top p, and system inputs. For testing these parameters, we used the final prompt design presented in Figure \ref{fig:Enhanced_prompt}.

\textbf{Temperature.} Temperature is a hyperparameter that allows controlling the randomness and creativity of the text generated by a GenAI. If the temperature is low, the model will probably produce the most “correct” text, but with little variation. Conversely, a higher temperature value shows greater variation (i.e., creativity). Lower temperature values are preferable for the development of a deterministic framework. We automated queries to ChatGPT to measure this feature, using the same prompt for all 33 privacy policies in the experimental dataset and performing the requests on 3 different days and at 5 different times of the day (from 9 am to 9 pm). This amounts to 495 different requests and responses, and we observed 52 discrepancies (i.e., different responses compared to the typical answer), which means 89.5\% consistency in ChatGPT responses. This —although far from absolute determinism— highly increases the 59.6\% percentage of determinism achieved with the default temperature value of 1.0.

The GPT-4 Turbo model introduced a new feature called seed, specifically for obtaining consistent responses over time with the same prompt. Even though determinism is its declared purpose, we observed that using a seed value and default temperature provided only 84.65\% of similar responses. Nonetheless, the combination of 0 temperature and seed shows 90.51\%, being the most reliable combination of these two parameters.

In our evaluation of the ChatGPT's performance across different temperature settings, we found that higher temperature values inversely impact the consistency of the metrics, with deterministic responses being optimal. This tendency is notable as the quality of the outputs deteriorates with increasing temperature. Concurrently, a manual inspection of the responses revealed a propensity for incomplete data type coverage. Specifically, responses frequently reported only the initial data type queried. This issue not only aggravates the decline in the F1 score but also results in a significant proportion of the data types—nearly half—remaining unaddressed in the responses. Such findings underscore the importance of temperature configuration in ensuring both the accuracy and completeness of the information extracted by ChatGPT.

\textbf{Top p.} Top p, or “\emph{nucleus sampling}”, consists of selecting the next token from the “nucleus” or subset of the vocabulary that constitutes the cumulative probability mass of the top p most probable tokens. For example, setting $top\_p=0.1$ means only tokens comprising the top 10\% probability mass are considered \cite{OpenAIAPI}. We have observed little performance variability when testing different values of $top\_p = $  $[0, 1]$ while keeping the default temperature value ($T=1$). Furthermore, in its official documentation, OpenAI recommends modifying the temperature value or the top p parameter, but not both simultaneously. Thus, we chose the default value of top p for our implementation and set the temperature to zero. These settings allow us to obtain more reproducible results.

\textbf{System inputs.} Using the OpenAI API, messages can be assigned to different roles (i.e., user, assistant, or system), where the system instruction can give high-level instructions for the conversation. We tested two system instructions: 1) “\emph{You are a helpful assistant with extensive knowledge in data protection and privacy engineering}.” and 2) “\emph{You are a helpful assistant with extensive knowledge in data protection and law}”, which specifically indicate areas of knowledge that are important for our task analyzing privacy policies. Neither of the two system instructions improved the results obtained but rather worsened them. 

\subsection{Fine-tuning}\label{Fine-tuning}


OpenAI facilitates model customization through fine-tuning, which involves re-training a model on a specific dataset to enhance its performance. This approach is beneficial for augmenting response consistency and can enable the use of shorter prompts while still achieving the desired format. During our experimentation phase, fine-tuning was available only for the \emph{gpt-3.5-turbo-0613} model. Thus, we tested the effect of model fine-tuning using this ChatGPT version.  

The \emph{gpt-3.5-turbo-0613} model sets a maximum prompt size of 4,096 tokens. Thus, we segmented the policies into smaller subsets (chunks), each conforming to the condition that the combined length of the policy text ($T$) and prompt ($P$) did not exceed 4,096 tokens ($T + P < 4,096$). This process led to the creation of a training set comprising 73 chunks, aligned with the manual annotations from the MAPP corpus and subjected to a default training configuration of three epochs as determined by OpenAI based on dataset size.

This fine-tuned model demonstrated superior performance when compared to the baseline (not fine-tuned) model: accuracy increased from 0.677 to 0.867, precision increased from 0.519 to 0.803, and the F1 score increased from 0.670 to 0.803. Still, it could not beat the GPT-4 Turbo model (which does not require chunking the policies thanks to the 128K tokens limit), probably due to its ability to retain context (see Table \ref{tab:fine_tuning_comparison}).

\begin{table}[h]
\caption{Comparison between the baseline and the fine-tuned GPT-3.5 model analyzing privacy policies by chunks and GPT-4 Turbo model analyzing privacy policies as a whole.}\label{tab:fine_tuning_comparison}
\centering 
\small 
\begin{tabular}{lccc}
\toprule
& \multicolumn{2}{c}{Chunked privacy policy processing} & \multicolumn{1}{c}{\parbox{35mm}{\centering Whole privacy policy processing}} \\
\cmidrule(lr){2-3} \cmidrule(lr){4-4}
Metrics & \parbox{25mm}{\centering gpt-3.5-turbo-\\0613 (fine-tuned)} & \parbox{25mm}{\centering gpt-3.5\\-turbo-0613} & \parbox{20mm}{\centering GPT-4 Turbo} \\
\midrule
Accuracy & 0.867 & 0.677 & 0.916 \\
Precision & 0.803 & 0.519 & 0.898 \\
Recall & 0.803 & 0.944 & 0.963 \\
F1 score & 0.803 & 0.670 & 0.935 \\
\bottomrule
\end{tabular}
\end{table}

\subsection{Validation}

We first applied our proposed configuration framework to the MAPP ground-truth control set, comprising 31 privacy policies, utilizing the prompt, parameters, and model based on our findings in prior sections. The prompt employed is the one described in Section \ref{Final_Prompt_Section}. The selection of parameter values was based on determinism consideration: $temperature=0$, a fixed seed, and $top\_p=1$ (the default setting). Finally, the GPT-4 Turbo model is employed for its performance, speed, and significantly higher input token limit balance. This configuration yields an accuracy of 0.916, a recall of 0.976, a precision of 0.898, and an F1 score of 0.935 on the control set of the MAPP corpus.

We further validated our prompt design and model configuration against a larger ground truth, i.e., OPP-115 dataset \cite{wilson2016websitePrivacyPolicy}, renowned for its fine-grained manual annotations of privacy practices. This validation yielded consistent results: 0.904 accuracy, 0.912 recall, 0.949 precision, and 0.930 F1 score, indicating that our proposal exhibits robust performance even when applied to a larger and more varied set of privacy policies.

\section{Demonstration}\label{Demonstration}

This section aims to demonstrate why LLMs, specifically ChatGPT, can be considered a competent technique for privacy policy analysis at scale. First, we compare ChatGPT to its closest GenAI rival, Llama 2, in terms of extracting the same privacy practices across identical test sets. We then compare ChatGPT with state-of-the-art statistical and symbolic NLP approaches to evaluate its performance and verify whether it can rival or even replace them. Finally, we analyze our proposal's generalization capabilities for identifying other privacy practices, namely the declaration of international transfers in privacy policies.

\subsection{Comparison with Llama 2}

Llama 2 \cite{touvron2023llama} is a family of open-source LLMs released by Meta that competes with ChatGPT in the GenAI space. Specifically, Meta has released versions with 7, 13, and 70 billion parameters, each with a fine-tuned “Chat” version optimized for dialogue. For a more direct comparison with the ChatGPT models, we focus on the chat variant of each of the Llama 2 models. 

We downloaded the 7B directly from Meta via their GitHub repository and ran it using four NVIDIA GeForce RTX 2080 Ti GPUs. Due to GPU constraints, we used the Python library from Together.AI to run the 70B model \cite{TogetherAIPythonDoc}. We initially tried to run the 13B model in our local environment, but as we achieved poor performance, we also used the Together.AI installation. 

\textbf{Prompt.} We departed from the final prompt design shown in Figure \ref{fig:Enhanced_prompt} and followed another split testing process to identify the best-performing Llama 2 prompt design. All the Llama 2-Chat models have a 4,096 token limit, which forced us to segment the privacy policies (i.e., the Data part) to ensure that our prompts are under the maximum token limit. Additionally, we removed the few-shot learning part from the prompt, as this yielded worse performance in our experiments with Llama 2. We observed that this technique resulted in outputs that did not conform to the requested format and additionally caused overfitting to the provided examples. Finally, we tried segmenting the Task in the prompt by asking for each data practice at a time, improving the results. Table \ref{tab:summary_tests} summarizes the different tests and the resulting performance.

\textbf{Parameters.} Just as for the ChatGPT models, we parameter-tuned across the temperature and the top p values. Similar to observations with ChatGPT, our experiments suggest setting a temperature value of 0 and a default top p of 1.0 as the best-performing configuration.   

We carried out our experiments with the three Llama 2 versions. As the Llama 2 70B-Chat model consistently showed better results, we used this version to assess its performance against the MAPP control set (31 annotated policies) (Table \ref{tab:Llama2_vs_ChatGPT}). Llama 2 demonstrates comparable but slightly lower performance in identifying privacy practices in this dataset compared to our ChatGPT-4 proposal. 

We further evaluated the performance of the Llama 2-70B configuration against the OPP-115 dataset. The results (Table \ref{tab:Llama2_vs_ChatGPT}) show that Llama 2 obtains worse performance against this new dataset, suggesting that, unlike ChatGPT, this Llama 2 configuration does not generalize well to different datasets.


\begin{table}[ht]
\centering
\caption{Metrics comparison between the best-performing Llama 2 70B-Chat (Llama 2) and ChatGPT-4 Turbo configurations.}\label{tab:Llama2_vs_ChatGPT}
\begin{tabular}{@{}clcccc@{}}
\toprule
 &  & \textbf{Accuracy} & \textbf{Precision} & \textbf{Recall} & \textbf{F1 score} \\ 
\cmidrule(lr){3-6}
\addlinespace[0.5em]
\multirow{2}{*}{\textbf{MAPP}} & Llama 2  & 0.846 & 0.847 & 0.921 & 0.882 \\
 & GPT-4 Turbo  & 0.916 & 0.898 & 0.976 & 0.935 \\
\addlinespace[0.5em]
\addlinespace[0.5em]
\multirow{2}{*}{\textbf{OPP-115}} & Llama 2  & 0.749 & 0.700 & 0.814 & 0.753 \\
 & GPT-4 Turbo  & 0.904 & 0.949 & 0.912 & 0.930 \\ 
\addlinespace[0.5em]
\bottomrule
\end{tabular}
\end{table}

\subsection{Comparison with state-of-the-art techniques}

We propose LLMs and, specifically, ChatGPT as a new technique for automating privacy policy information processing and extraction. To confirm it as such, we compare its performance in extracting fine-grained practices from policies with state-of-the-art statistical and symbolic approaches.

\subsubsection{Statistical approaches}

In this study, we conducted a comparative analysis of our configuration framework proficiency in identifying fine-grained privacy practices against statistical classifiers based on Support Vector Classifiers (SVC) —a subtype of SVM—, which were trained and validated using the APP-350 corpus \cite{zimmeck2019maps}. To ensure a rigorous comparison, the same policy dataset was employed to evaluate the performance of both methods.

The primary objective was to assess ChatGPT’s ability to accurately identify particular types of personal data collection as stated in privacy policies. For this purpose, we selected 10 distinct data types, with an emphasis on higher specificity (for instance, choosing “Contact email address” over the broader “Contact information”), spanning various categories such as contact data, identifiers, and social login data.

Table \ref{tab:ML_vs_ChatGPT} delineates the comparative performance of ChatGPT against the pre-trained SVC classifiers for identifying each specified data type. The results indicate a comparable level of performance across most data types. However, a notable exception was observed with the SIM identifier, where ChatGPT's performance was significantly lower despite achieving 100\% precision. A detailed manual review of the annotations for this data type in the original policies revealed a common annotation issue: Human annotators wrongly coded this data type. Specifically, the annotators coded “device serial number” under the “SIM serial number” category. However, the former is issued by the device manufacturer, while the latter is provided by the mobile carrier. This discrepancy likely contributed to the lower F1 score for ChatGPT in identifying the SIM identifier.

\begin{table}[ht]
\centering
\caption{ChatGPT vs. traditional machine learning classifiers' performance for identifying first-party data collection per data type.}\label{tab:ML_vs_ChatGPT}
\begin{tabular}{lcccccc}
\toprule
& \multicolumn{3}{c}{\textbf{ChatGPT}} & \multicolumn{3}{c}{\textbf{SVC classifiers}} \\
\cmidrule(lr){2-4} \cmidrule(lr){5-7}
\textbf{Data type} & \textbf{Precision} & \textbf{Recall} & \textbf{F1 score} & \textbf{Precision} & \textbf{Recall} & \textbf{F1 score} \\
\midrule
Contact Email Address & 95\% & 95\% & 95\% & 97\% & 94\% & 96\% \\
Contact Phone Number & 100\% & 85\% & 92\% & 94\% & 94\% & 94\% \\
Identifier Cookie & 92\% & 97\% & 95\% & 95\% & 100\% & 98\% \\
Identifier IMEI & 83\% & 88\% & 86\% & 94\% & 94\% & 94\% \\
Identifier Device ID & 74\% & 89\% & 81\% & 96\% & 87\% & 91\% \\
Identifier MAC & 94\% & 84\% & 89\% & 88\% & 79\% & 83\% \\
Identifier Mobile Carrier & 79\% & 90\% & 84\% & 100\% & 57\% & 73\% \\
Identifier SIM Serial & 100\% & 13\% & \textbf{22\%} & 73\% & 100\% & 84\% \\
Location WiFi & 70\% & 58\% & 64\% & 48\% & 92\% & 63\% \\
Social login & 77\% & 65\% & 71\% & 83\% & 81\% & 82\% \\
\bottomrule
\end{tabular}
\end{table}

Excluding the analysis of the SIM serial identifier, which was identified as an anomaly, the comparative evaluation yielded an average F1 score of 84.1\% for ChatGPT, as opposed to 86\% achieved by the SVC-based classifiers for the selected data types. This outcome illustrates that while traditional SVC-based classifiers are recognized for their reliability and accuracy, ChatGPT presents a comparable level of performance. ChatGPT offers the added advantage of significantly simpler usability, making it a viable alternative for similar tasks in data practice identification.

\subsubsection{Symbolic approaches}

PolicyLint \cite{Andow2019PolicyLint}, a tool designed to analyze privacy policies, employs a symbolic approach based on ontologies to detect contradictions in statements regarding personal data collection and sharing. This tool identifies negative sentences, which are often challenging for conventional machine learning techniques. The public repository of PolicyLint's code \cite{PrivacyPolicyAnalysis}, as referenced, was utilized to process the privacy policies in our control set, facilitating a comparative analysis with our proposal.

PolicyLint operates by identifying sentence structures characterized by [actor] [action] [data\_object] [entity]. Here, “actor” signifies a first or third party involved in data handling, “action” denotes the nature of data interaction (positive or negative, such as collection or non-collection), “data\_object” pertains to the type of data in question, and “entity” refers to the recipient of the data (for instance, advertisers).

\begin{figure}[H]
  \centering
  \includegraphics[width=0.75\linewidth]{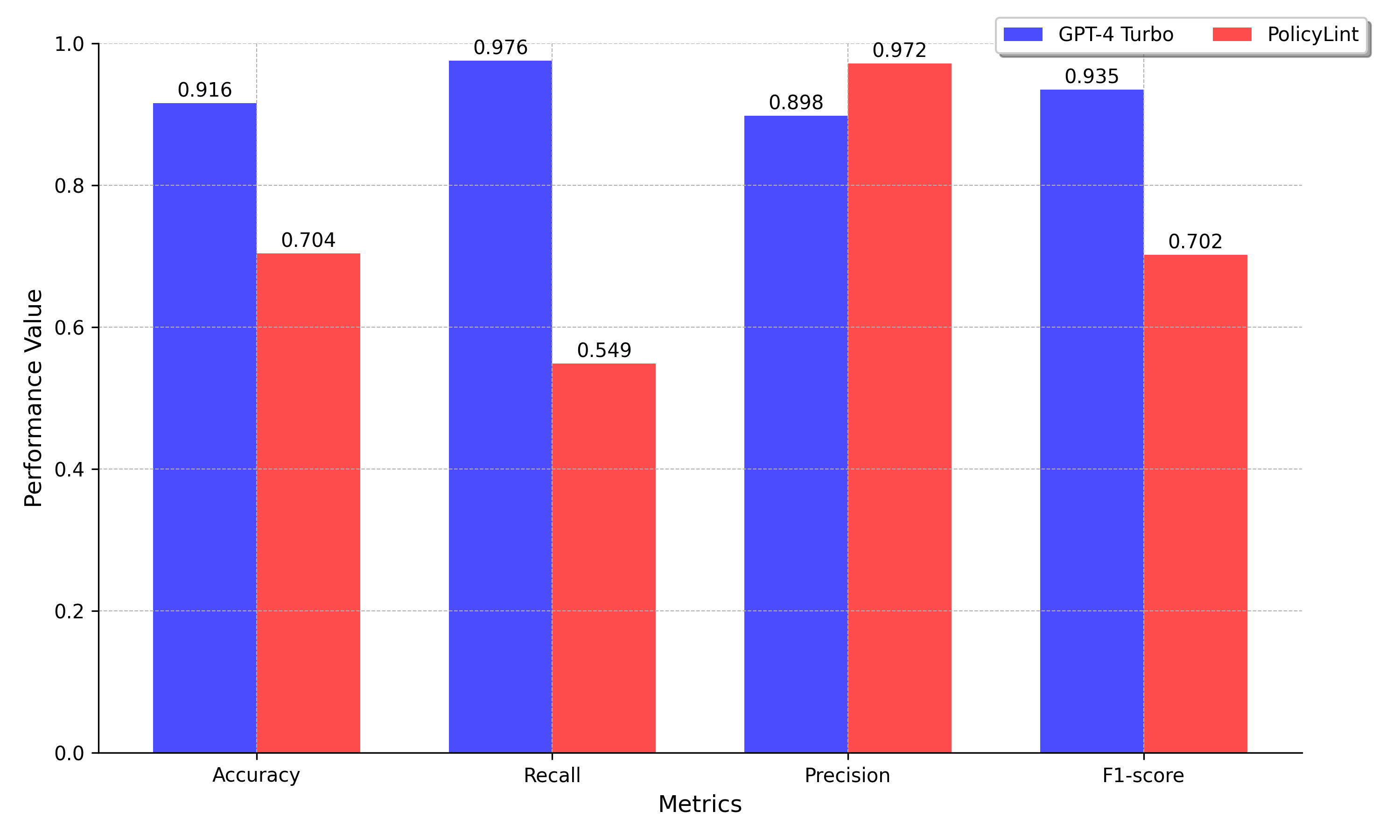}
  \caption{Metrics comparison between the ChatGPT-based method and PolicyLint.}
  \label{fig:PolicyLint_comparison}  
\end{figure}

Given that the data\_objects in PolicyLint do not align format-wise with those in our MAPP corpus, a manual matching process was undertaken by two authors to correlate PolicyLint's classifications with the data types in our corpus. This matching was independently conducted, followed by an agreement phase for resolving discrepancies. The matching criteria were aligned with the definitions provided in the MAPP corpus. Subsequently, the comparative performance metrics of both methods were analyzed and presented in Figure \ref{fig:PolicyLint_comparison}.

Our analysis revealed that PolicyLint exhibits high precision, surpassing the metrics its authors reported. This discrepancy might stem from our methodology, where we assess whether a data type is identified at least once in a policy, instead of PolicyLint's validation across all relevant statements. However, PolicyLint's approach overlooks negative cases, leading to a lower recall. Overall, our evaluation indicates that our ChatGPT configuration framework significantly outperforms PolicyLint's F1 score, highlighting its efficacy in extracting and analyzing data practices from privacy policies.

\subsection{Generalization Capabilities}

The proficiency of ChatGPT in extracting data collection and sharing practices from privacy policies has been notably demonstrated in our study. In this section, we extend the evaluation to assess ChatGPT's generalization capabilities in identifying a broader range of practices within privacy policies. This extension is grounded in our prior research \cite{Guaman2023}, which focused on analyzing privacy policies to find disclosures related to international data transfers. This previous study produced a dataset (IT100) comprising 100 privacy policies where privacy practices related to international data transfers were manually annotated by legal experts \cite{IT100Corpus}. A Support Vector Machines (SVM)-based classifier was trained to identify these specific practices.

\begin{figure}[H]
  \centering
  \includegraphics[width=0.75\linewidth]{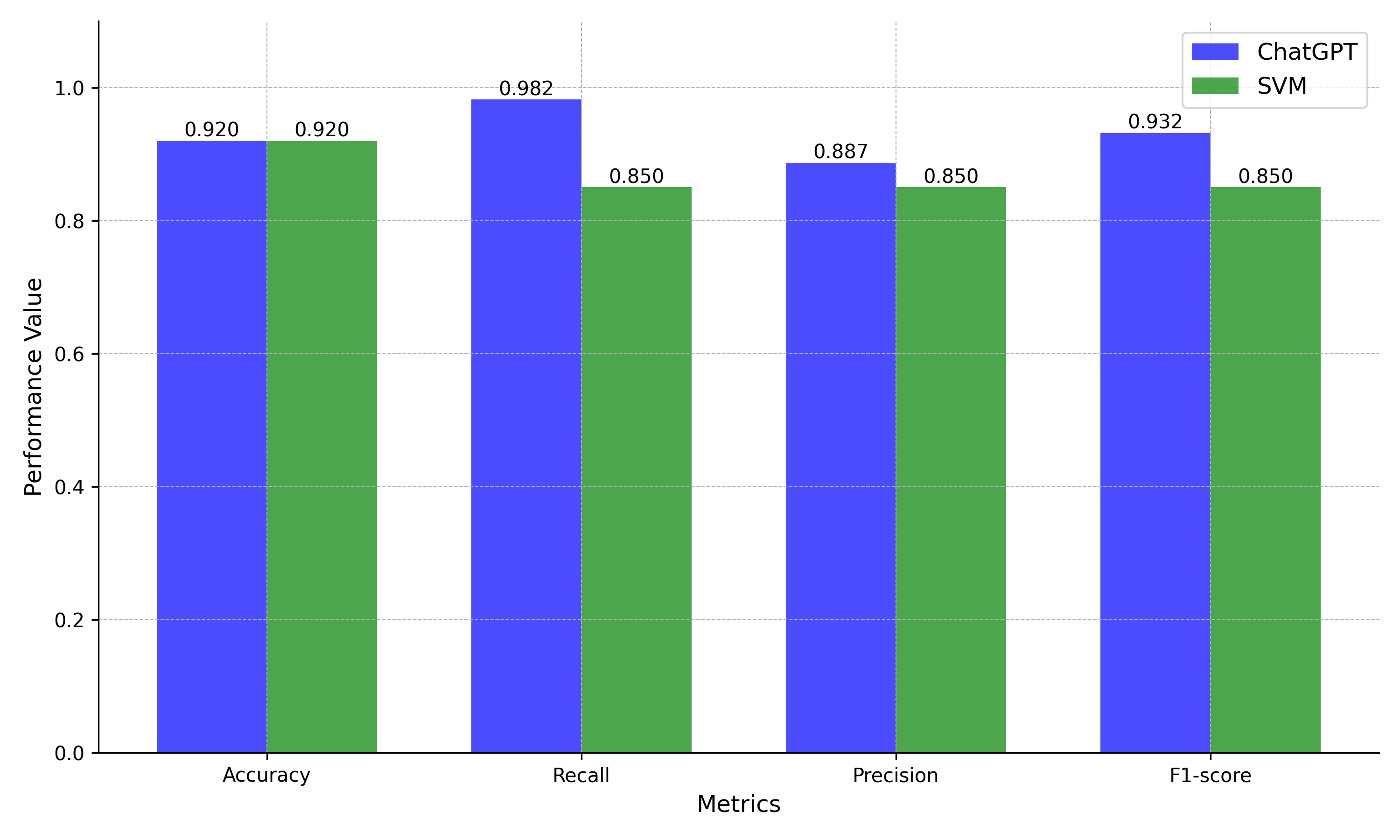}
  \caption{Metrics comparison between the ChatGPT and SVM machine learning classifier performance identifying international data transfer practices.}
  \label{fig:Int_Transfers_comparison}  
\end{figure}

In Figure \ref{fig:Int_Transfers_comparison}, we present the comparative analysis of the performance metrics between our configuration proposal of ChatGPT and the SVM-based classifier, utilizing the IT100 dataset for evaluation. ChatGPT was configured as per the parameters delineated in Section \ref{Experimental Design}, which included an instantiation of our enhanced prompt in Section \ref{Final_Prompt_Section}, and settings like temperature=0, top\_p=1, and an absence of system\_input. The results displayed by ChatGPT were significantly superior in most metrics, reinforcing its efficacy in extracting information about diverse practices from privacy policies.

\section{Discussion}

\textbf{ChatGPT demonstrates a more balanced and adaptable performance in privacy policy analysis compared to traditional symbolic and statistical methods, overcoming the limitations of manual annotations and varying data across different corpora}. Symbolic methods are characterized by their rigidity, which is reflected in their performance metrics. High precision in symbolic methods indicates their well-defined patterns and rules are closely aligned with specific instances in the data. However, this precision comes at the cost of completeness, as evidenced by their lower recall. In contrast, ChatGPT demonstrates a more balanced performance, achieving a notably higher F1 score than PolicyLint. This suggests that ChatGPT –and our proposed configuration–, while less rigid in its approach, captures the breadth of privacy practices within policies more effectively.

When comparing ChatGPT with statistical methods such as SVM, we find that these traditional classifiers perform similarly in identifying certain data types. However, ChatGPT excels particularly in recognizing practices like international data transfers, which are complex and multifaceted. This superior performance is notable, given that statistical methods often depend on extensive manual annotations, which can introduce errors. As seen in the Identifier SIM Serial case in Table \ref{tab:ML_vs_ChatGPT}, such annotation errors can significantly impact classifier performance. Wagner et al. \cite{Wagner2023} supports this observation, indicating that the average agreement among human annotators for attribute values is considerably lower than for top-level categories. This discrepancy highlights the challenges in achieving consensus among annotators and the advantage of ChatGPT's approach, which is not constrained by the limitations of manual annotations.

Furthermore, our analysis of different corpora, specifically the MAPP and OPP-115 datasets, sheds light on the variance in annotations across datasets. The performance disparities observed for \emph{Social media data} and \emph{Personal identifier data} between these two corpora suggest that the annotations for these data types likely vary, underscoring the issues associated with training classifiers on manually annotated data \cite{yan2014learning}. This reinforces the need for approaches like ChatGPT that rely less on such annotations, offering a more adaptable and potentially more accurate solution for privacy policy analysis.

\textbf{Economic considerations play a significant role in the choice of the technique to process privacy policies}. Manual annotators in the United States are reported to earn approximately \$8.5 per hour, while rates in lower-income countries range between \$3 to \$4 per hour \cite{AI2Crowdsourcing}. However, the annotation of privacy policies demands legal expertise for accurately identifying data protection practices, entailing a higher pay rate, assumed here at a minimum of \$10 per hour. For the MAPP corpus, three experts annotated each policy, averaging 1 hour and 52 minutes each \cite{arora2022privacyPolicyCorpus}. Multiple annotations of the same content by different experts ensure reliable and high-quality data where inter-annotation agreement can be measured. Previous research \cite{Guaman2023} demonstrated that training classifiers with 100 policies can be sufficient, which raises costs by up to \$5,601.

Setting aside the technical expertise required for classifier development, the cost-effectiveness of traditional classifiers becomes behooveful with GPT-4 Turbo at approximately 81,500 privacy policies and with GPT-3.5 Turbo at around 825,000 policies. This cost difference suggests that depending on specific application needs and constraints, GPT-3.5 Turbo, with an F1 score of 87.2\% measured on the MAPP corpus control set, might be a pragmatic choice compared to GPT-4 Turbo, which achieved an F1 score of 93.5\%\footnote{This difference is not only due to the model performance but also because the few-shot prompting technique that can be applied to the GPT-4 Turbo model thanks to its increased token limit.} in our evaluation. Furthermore, Llama 2 models, specifically 7B and 70B, may be considered in terms of cost discussion. Both models were publicly released for free use, but our hardware limitations imposed by the latter forced us to use Together.AI API for that version. The current API cost for the Llama 2-70B model is 10\% lower than the GPT-3.5 Turbo model while showing an even higher performance –88.2\% F1 score– making it even more convenient in terms of cost by performance. Llama 2-7B has significantly lower computational requirements, leading to no other cost but computation and achieving an 80.1\% F1 score. Thus, GPT-4 Turbo offers the best performance of the LLMs compared, but at the highest cost. Whereas if the computing capabilities are sufficient to run it locally, Llama 2-70B offers good performance at a low cost.

\begin{figure}[H]
  \centering
  \includegraphics[width=0.75\linewidth]{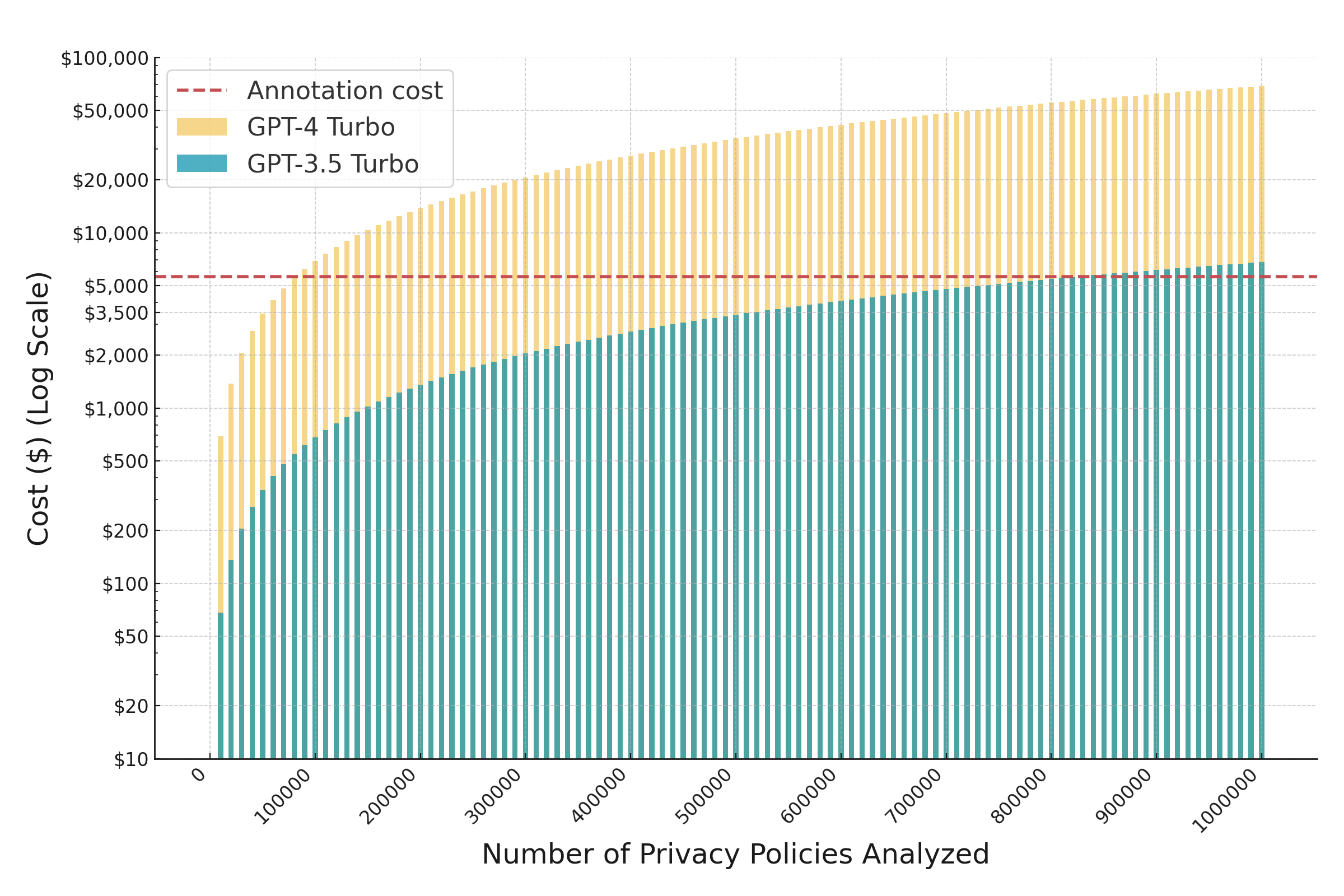}
  \caption{Cost Comparison of analyzing privacy policies with ChatGPT API.}
  \label{fig:ChatGPT_api_costs}  
\end{figure}

\textbf{In assessing the processing capabilities of ChatGPT models, our analysis indicates a marked efficiency advantage over traditional machine learning and symbolic AI techniques}. Acknowledging the operational constraints imposed by OpenAI on these models, specifically regarding token throughput per minute is critical. GPT-4 Turbo is limited to 300,000 tokens per minute, while GPT-3.5 Turbo can process up to 1,000,000 tokens within the same timeframe.

\begin{figure}[H]
  \centering
  \includegraphics[width=0.75\linewidth]{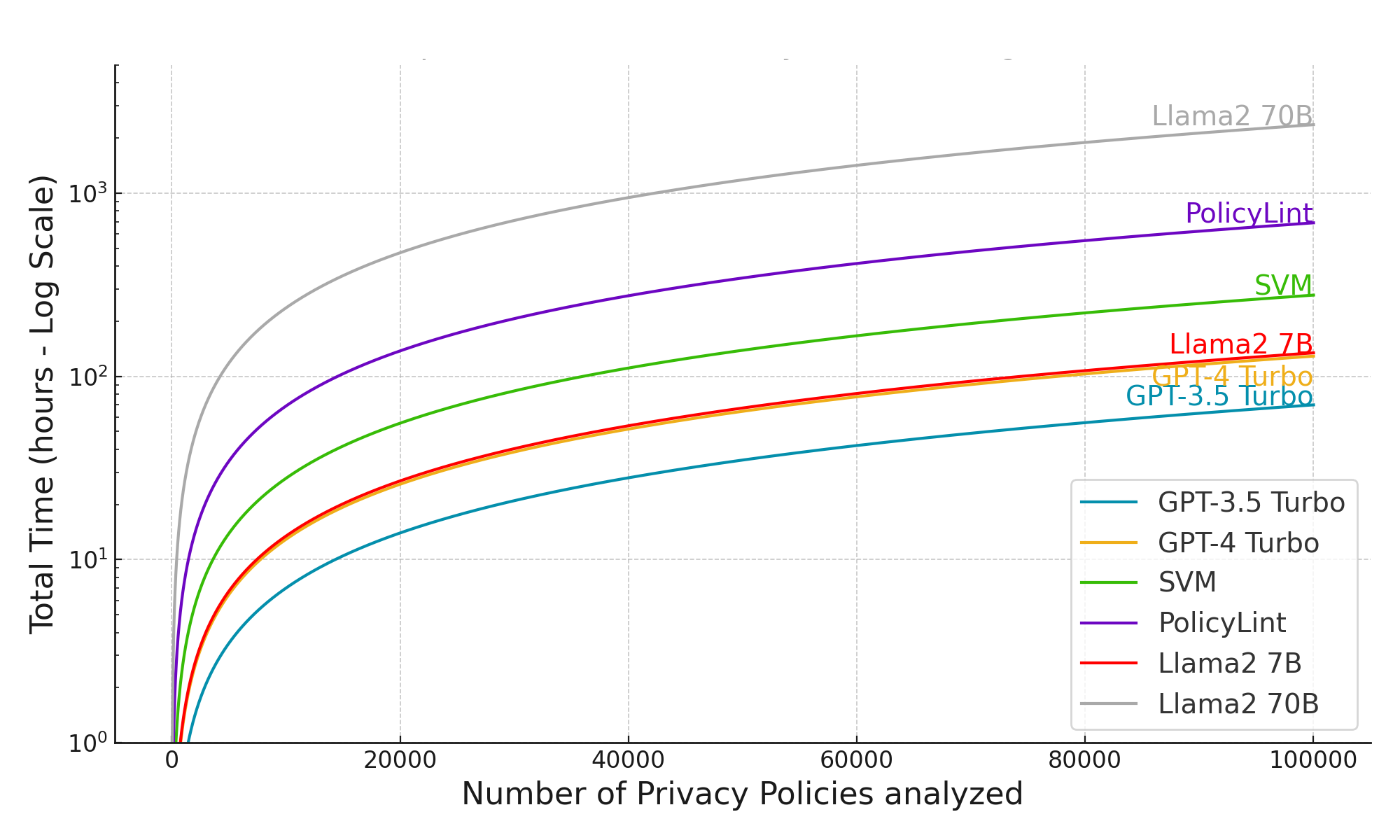}
  \caption{Time comparison between ChatGPT models and SVM to process privacy policies.}
  \label{fig:ChatGPT_api_time}  
\end{figure}

With an average of 6,652 tokens required in average to fully process a privacy policy, GPT-4 Turbo can analyze up to 45 policies per minute, in contrast to the 150 policies per minute capability of GPT-3.5 Turbo. Figure \ref{fig:ChatGPT_api_time} depicts this variance in processing capacity, with GPT-4 Turbo necessitating slightly more time for large-scale privacy policy analyses when compared to GPT-3.5 Turbo. Furthermore, the SVM-based classifier takes approximately double the time of the slower GPT model to process an equivalent number of policies. In stark contrast, PolicyLint, while being at the forefront of privacy policy analysis symbolic-based tools, demands up to six times the processing time of GPT-3.5 Turbo for comparable tasks. The two versions of Llama 2 show remarkably different processing times. The Llama 2-7B, locally analyzing each policy at once (truncating policies when length limit required), shows a similar processing time compared to GPT-4 Turbo, while Llama 2-70B through Together.AI API (analyzing policies by chunks), shows the slowest performance of all techniques.

These findings underscore the superior speed of LLM models and highlight the need to balance performance with processing time, especially when scaling to analyze vast numbers of privacy policies. Thus, organizations may find the trade-off between the slightly lower speed of GPT-4 Turbo and its enhanced accuracy acceptable, particularly in scenarios where quality of analysis is paramount. Conversely, for applications where time efficiency is a priority, GPT-3.5 Turbo presents a compelling option, offering rapid analysis with a modest compromise in performance metrics.

For the GPT models, parallel processing can be employed to concurrently analyze up to 150 and 45 policies per minute for GPT-3.5 and GPT-4, respectively, adhering to the stipulated token rate limits. To scale up concurrent processing capabilities with ChatGPT, users may opt for multiple paid accounts, which entails additional costs due to the subscription requirements for accessing the API via ChatGPT Plus. Another avenue is to request OpenAI for elevated rate limits, a request that hinges on the company’s approval. Anticipation of expanded rate limits by OpenAI in the future could potentially democratize access to more extensive parallel processing for all users, thereby broadening –even more– the scope of large-scale privacy policy analysis.

\textbf{The rapid progression of generative AI technology is evident in the quick succession of ChatGPT models introduced}. Within the span of mere months, we have witnessed the release of successive ChatGPT iterations, namely GPT-3.5 Turbo, GPT-4, and GPT-4 Turbo. Alongside the expected speed and cost efficiency enhancements, a notable shift has been observed in model determinism. For instance, the determinism observed in ChatGPT-3.5 (99.19\%) significantly exceeds that of GPT-4 Turbo, suggesting a potential trade-off between response variability and model robustness.

This rapid succession has introduced variations in the models' performance, particularly regarding prompt responsiveness and temperature settings. Current outputs from most recent models align more closely with expectations even at increased temperature settings, evidencing an enhanced capacity of ChatGPT to interpret prompts with fewer instructions and diminishing the necessity for techniques such as prompt augmentation.

The execution speeds of the Turbo models are noteworthy, achieving significant throughput without compromising performance for the task at hand. Moreover, the cost efficiencies introduced with these models —threefold less for GPT-4 Turbo and tenfold less for GPT-3.5 Turbo— consolidate ChatGPT's position as a vying competitor to state-of-the-art tools for large-scale studies.

We have observed that the token limit per minute has substantially increased—up to 30 times for GPT-4 and nearly 10 times for GPT-3.5 Turbo. This escalation, coupled with the models' improved response times, results in more expedient processing of privacy policies, as evidenced in Figure \ref{fig:ChatGPT_api_time}. Regarding F1 score performance, the new GPT-4 Turbo model remains consistent with its predecessors, albeit with notable variations: a 1.36\% increase in the F1 score for the MAPP corpus and a similar decrease for the OPP-115. The intricacies of these models make it challenging to pinpoint the exact causes of these variations, but it is remarkable that the optimization inherent in the Turbo models has not detrimentally impacted performance for this specific task.

\section{Conclusion}




Throughout this article, we have substantiated the applicability of LLMs in analyzing and extracting privacy practices from privacy policies. Specifically, ChatGPT has proven to be as effective as traditional NLP techniques, offering significant advantages in terms of cost, runtime, and ease of development. This work has also presented a tailored configuration of prompts, parameters, and the ChatGPT model, which shows outstanding performance in identifying various privacy practices within privacy policies. We have demonstrated that fine-tuning, while valuable, may not be the most optimal approach against the backdrop of few-shot and zero-shot learning paradigms. Intriguingly, few-shot learning has exhibited superior performance metrics even over zero-shot learning.

Our study demonstrates a notable advance in automated privacy policy analysis through the generalization capabilities of our proposal. This approach potentially eliminates the reliance on annotated datasets, enabling the analysis to encompass a wider array of privacy practices previously limited by the necessity for in-depth legal knowledge. Future work will focus on integrating this method into automated systems for assessing data protection compliance, contributing towards raising awareness among developers, regulators, and users of the potential privacy risks in the data protection ecosystem.

\section{Aknowledgments}
This work has been partially supported by the TED2021-130455A-I00 project funded by MCIN/AEI/10.13039/501100011033 and by the European Union “NextGenerationEU”/PRTR. Jose M. del Alamo has received a grant from the Spanish “Ministerio de Universidades” through the “Movilidad” sub-programme of the “Programa Estatal para Desarrollar, Atraer y Retener Talento”, within the “Plan Estatal de Investigación Científica, Técnica y de Innovación 2021-2023”. This research has also been partially supported by the National Science Foundation under its Security and Trustworthy Computing Program (grant CNS-1914486).

\bibliographystyle{unsrt}  
\bibliography{templateArxiv}

\end{document}